\documentclass[10pt, conference, compsocconf]{IEEEtran}
\usepackage{wrapfig}
\usepackage{mathtools}
\usepackage{autonum}
\usepackage{tabularx}
\newcolumntype{R}{>{\raggedleft\arraybackslash}X}

\newcommand*{\maximize}{\operatornamewithlimits{maximize}}
\newcommand*{\sov}{\operatornamewithlimits{SOV}}
\newcommand*{\pppa}{\operatornamewithlimits{PPPA}}
\newcommand*{\auroc}{\operatornamewithlimits{AUC}}
\newcommand*{\minov}{\operatornamewithlimits{minov}}
\newcommand*{\maxov}{\operatornamewithlimits{maxov}}
\newcommand*{\len}{\operatornamewithlimits{len}}
\newcommand*{\ef}{\operatornamewithlimits{F1}}
\newcommand*{\score}{\operatorname{score}}
\ifCLASSINFOpdf
   \usepackage[pdftex]{graphicx}
\else
\fi
\usepackage[font=footnotesize]{subfig}
\begin{document}
%
\title{Text segmentation on multilabel documents: A distant-supervised approach}


\author{\IEEEauthorblockN{Saurav Manchanda}
\IEEEauthorblockA{University of Minnesota \\
Twin Cities, MN 55455, USA \\
manch043@umn.edu}

\and

\IEEEauthorblockN{George Karypis}
\IEEEauthorblockA{University of Minnesota \\
Twin Cities, MN 55455, USA \\
karypis@umn.edu}
}


%


\maketitle

\begin{abstract}
Segmenting text into semantically coherent segments is an important task with applications in information retrieval and text summarization. Developing accurate topical segmentation requires the availability of training data with ground truth information at the segment level. However, generating such labeled datasets, especially for applications in which the meaning of the labels is user-defined, is expensive and time-consuming. In this paper, we develop an approach that instead of using segment-level ground truth information, it instead uses the set of labels that are associated with a document and are easier to obtain as the training data essentially corresponds to a multilabel dataset. Our method, which can be thought of as an instance of distant supervision, improves upon the previous approaches by exploiting the fact that consecutive sentences in a document tend to talk about the same topic, and hence, probably belong to the same class. Experiments on the text segmentation task on a variety of datasets show that the segmentation produced by our method beats the competing approaches on four out of five datasets and performs at par on the fifth dataset. On the multilabel text classification task, our method performs at par with the competing approaches, while requiring significantly less time to estimate than the competing approaches. 

\end{abstract}

\begin{IEEEkeywords}
distant supervision; multilabel; segmentation;

\end{IEEEkeywords}

%
\IEEEpeerreviewmaketitle

\section{Introduction}
\begin{figure}[!t]
\includegraphics[width=\columnwidth]{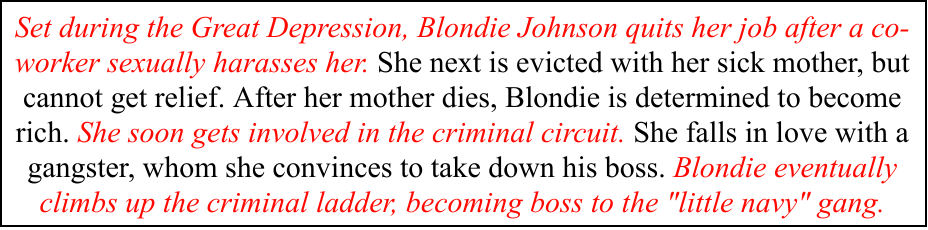}
\caption{Plot summary of the movie \textit{Blondie Johnson}. Red, italicized text corresponds to the crime genre and black, unitalicized text belongs to the drama genre.}
\label{motivation_movie}
\end{figure}

A document can be visualized as a sequence of semantically-coherent segments. Text segmentation refers to the task of breaking down the documents into these segments. Segmenting the text into these semantically-coherent segments is useful in many natural language processing tasks~\cite{hearst1997texttiling}: it can improve information retrieval (by indexing/recognizing documents more precisely or by giving the specific part of a document corresponding to the query as a result); and text summarization (by including information from each of the classes present in a document). In the case of multilabel documents, these segments can be mapped to one or more of the labels associated with the document. For example, Fig. \ref{motivation_movie} shows the plot summary of the 1933 movie \textit{Blondie Johnson}. The movie belongs to the crime/drama genre. We can segment the plot summary text into segments belonging individually to the crime genre and drama genre (Red, italicized text corresponds to the crime genre and black, unitalicized text belongs to the drama genre). Manually generating these segments and annotating them with the labels in a document is a tedious and expensive task. The task of associating individual parts in a document with their most appropriate labels is termed as \textit{credit attribution}~\cite{ramage2009labeled}.

In this paper, we present a method to accurately and efficiently solve the \textit{credit attribution} problem. Our method exploits the structure within the documents by leveraging the fact that the sentences occurring together in a document have high chances of belonging to the same class. Our method employs an iterative approach. In each iteration, it estimates a model for predicting the classes of an arbitrary length text segment, then uses dynamic programming to segment the document into semantically-coherent chunks using the estimated model, and fine-tunes the model with these segments. Every iteration provides a more precise classification model and hence, better segmentation. We evaluated the performance of our model on five datasets belonging to varying domains. On the segmentation task, our model performs better than Multi-Label Topic Model (MLTM)~\cite{soleimani2017semisupervised} on four out of five datasets and comparable performance on the fifth dataset, with the performance gain of $7.7\%$ over MLTM, averaged over all the datasets. Moreover, our method takes significantly less time to estimate as compared to the MLTM. For example, on one of the datasets, our model takes as less as 26 minutes to estimate as compared to more than 19 hours taken by MLTM. In addition, with segmentation, we have precise sections of text belonging to each class, which can help us to build more precise classification models. Therefore, we also evaluate our model on the multilabel text classification task, on which our model performs at par with the MLTM.

\section{Related Work} \label{literature}
Many methods have been developed to address the problem of text segmentation. The methods presented in this section associate individual words/sentences in a document with their most appropriate topic labels.

Labeled Latent Dirichlet Allocation (LLDA)~\cite{ramage2009labeled} is a probabilistic graphical model for \textit{credit attribution}. It assumes a one-to-one mapping between the class labels and the topics. 
Unlike LDA, LLDA incorporates supervision by constraining the topic model to use only those topics that correspond to a document’s labels. 
Partially Labeled Dirichlet Allocation (PLDA)~\cite{ramage2011partially} is a further extension of the LLDA that allows more than one topic for every class label, as well as some general topics that are not associated with any class. 

Multi-Label Topic Model (MLTM)~\cite{soleimani2017semisupervised} improves upon PLDA by allowing each topic to belong to multiple, one, or even zero classes probabilistically. MLTM also assigns a label to each sentence and the labels of the documents are generated from the labels of the constituent sentences. 

A common limitation of the above approaches is that they model the document as a bag of words/sentences and do not take into consideration the structure within the documents. Moreover, these methods are based on top of the LDA, whose inference is computationally expensive.

\section{Definitions, Notations and Background} \label{definitions}
Let $C$ be a set of labels and $D$ be a set of multilabel documents. For each document $d\in D$, let $L_d \subseteq C$ be its set of labels and let $d_q$ be the number of sentences that it contains. The approaches developed in this paper assume that in multilabel documents, consecutive subsets of sentences describe a single topic and that the same topic can be described by multiple such consecutive subsets of sentences. In particular, given a document $d$ we assume that there is a set of indices $0 = s_{0} < s_{1} < \ldots < s_{d_u} = d_q$ which partitions $d$ into $d_u$ non-overlapping contiguous segments:
\begin{equation}
S_d = \langle d[s_{0} + 1, \ldots, s_{1}], \ldots, d[s_{d_u-1} + 1, \ldots, s_{d_u}]\rangle,
\end{equation}
that span the entire document. The indices $s_{0}, \ldots, s_{d_u}$ correspond to changes in the topic being discussed in $d$. We refer to $S_d$ as the \emph{segmentation} of $d$. Let $H_{d}$ be the sequence of labels corresponding to the segmentation $S_d$, that is, $i$th element of $H_{d}$ ($ H_{di}$) is the annotated class label of the segment $d[s_{i} + 1,\ldots, s_{i+1}]$. Furthermore, we will use $d[i,\ldots, j]$ to refer to the part of the document starting at the $i$th sentence and ending at the $j$th sentence (inclusive), and we will refer to that part as the subsequence of $d$.

\section{Proposed method}\label{proposed}

Given a document $d$ and its associated set of labels $L_d$, we seek a segmentation $S_d$ and the corresponding segment-labels $H_d$ ($H_d \subseteq L_d$) such that each sentence in the segment $S_{di} \in S_d$ belongs to the class $H_{di}$. We can quantify the degree to which a text segment belongs to a class by using a multiclass classifier, which takes a piece of text of arbitrary length as input and gives as output the likelihood of that text belonging to each of the classes in $L_{d}$. We intend to find a segmentation that maximizes the aggregated score (likelihood) of all the segments belonging to their respective classes. This leads to an optimization function of the form:

\begin{equation}\label{eq:objective_function}
    J(d) = \maximize_{S_d, H_d} \sum_{i = 0}^{d_u - 1}\score(d[s_{i} + 1,\ldots,  s_{i+1}], H_{di}),
\end{equation}

\noindent where $\score(d[s_{i} + 1,\ldots,  s_{i+1}], H_{di})$ is the likelihood that the segment $d[s_{i} + 1,\ldots,  s_{i+1}]$ belongs to class $H_{di}$. 

According to Equation \ref{eq:objective_function}, in order to obtain a segmentation of $d$, we require two components: (i) A model for predicting the classes, which takes a piece of text of arbitrary length as input and gives as output the likelihood of that text belonging to each of the classes in $L_{d}$. (ii) An optimization algorithm for the objective function \ref{eq:objective_function}.

We explain each of the above two parts in the next two sections. First, we look into the segmentation algorithm, and then we look into how to build the prediction model to get the likelihood of a text segment belonging to each of the classes. 

\subsection{Segmentation algorithm}
Our model computes a segmentation $S_d$ and assignment of labels to each segment. Given the additive formulation of our objective function (Equation \ref{eq:objective_function}), we use dynamic programming (DP) to obtain an efficient and optimal solution.
Let $F(d, i, j, l, R)$ be the objective value associated with optimal segmentation of the subsequence $d[i,\ldots, j]$ of $d$ such that the label associated with the first segment in the subsequence $d[i,\ldots, j]$ is $l$ and the labels associated with all segments (including $l$) are drawn from the set $R$. Using this definition, the objective value of the optimal segmentation for $d$, according to Equation \ref{eq:objective_function}, is given by: 
\begin{equation}\label{eq:overall}
    J(d) = \max_{l \in L_{d}}F(d, 1, d_q, l, L_{d}).
\end{equation}

In order to prevent the segmentation algorithm from creating very small segments that switch topics, we use the $\alpha$ parameter, which assigns a fixed cost to the creation of a new segment. This is motivated by the gap-opening costs, often used to obtain meaningful alignments in biological sequences~\cite{gusfield1997algorithms}.
The recurrence relation for $F(d, i, j, l, R)$ is:
\begin{multline}\label{eq:simple_segmentation}
    F(d, i, j, l, R) = \max_{l' \in R, l'\neq l, i<k\leq j}\\(\score(d[i,\ldots,  k], l) + F(d, k+1, j, l', R) - \alpha).
\end{multline}
The first term in Equation \ref{eq:simple_segmentation} corresponds to mapping a prefix of $d[i,\ldots, j]$ to label $l$. We discuss the $\score$ function in the next section. The second term corresponds to computing the optimal segmentation of the remaining suffix of $d[i,\ldots, j]$, whose first segment is not mapped to label $l$. Equation \ref{eq:simple_segmentation} decomposes the solution by considering all possible prefixes that can be mapped to label $l$ and choosing the one with the maximum objective value. Moreover, given an optimal solution such that $d[i,\ldots, k]$ is assigned to label $l$ and its $(k+1)$st element is assigned to a label $l'$, it can be shown that $F(d, k+1, j, l', R)$ is the value of the optimal solution of decomposing $d[k+1,\ldots, j]$. Hence, Equation \ref{eq:simple_segmentation} leads to the optimal segmentation.

\subsection{Building the classification model}
For the segmentation algorithm, we require a model that gives us the score corresponding to the likelihood of a text segment belonging to a class; i.e., the $\score$ function in Equation \ref{eq:simple_segmentation}. Our method achieves this by using a multiclass classifier that takes as input a text segment and outputs the likelihood of it belonging to each of the classes. 

Arguably, the most obvious way to train this classifier is by using each document-label pair as a training example, that is, we convert the multilabel documents into single-label documents by pairing every document to each of its associated labels independently. However, as discussed in the introduction, the labels of the document do not apply with equal specificity across the complete document. In this section, we look at a  method to improve over the training exercise by removing noise from these training examples. 

We developed an iterative method to train the classifier as follows: We start with estimating an initial classification model using the noisy examples in the simple manner described above. In each iteration, we perform segmentation on the training documents to get specific segments belonging to each class and fine-tune the classifier with these specific training examples. Therefore, in each iteration, we expect to get a better segmentation that we use to fine tune the classifier to make it more precise.

Thus, we have two approaches for computing the $\score$ function. The first estimates a noisy classifier by using each document-label pair as the training set. We call this approach \emph{segmentation with noise} (SEG-NOISE). The second refines the classifier iteratively by performing segmentation on the training documents in each iteration.  We call this approach \emph{segmentation with refinement} (SEG-REFINE).

We use a simple two-layer perceptron as our multiclass classifier. The input layer has the number of nodes equal to the vocabulary size ($|V|$), followed by a hidden layer, and an output layer of the size ($|C| + 1$). We add an extra node in the output node to capture the segments that do not belong to any of the classes (we call it the null class).  We apply Rectified Linear Unit (ReLU) on the hidden layer and softmax on the output layer. We minimize the cross entropy loss for training the model. For the SEG-REFINE, we find the text segments belonging to each class (except the null class) and update the network with a smaller learning rate.

\section{Experimental methodology}\label{experiments}
\subsection{Datasets}
We performed experiments on five multilabel text datasets belonging to different domains as described below:
\par \emph{Movies}~\cite{bamman2014learning}: This dataset contains movie summaries and their genres. We randomly take a subset of the movies from this dataset corresponding to six common genres: \emph{Romance Film}, \emph{Comedy}, \emph{Action}, \emph{Thriller}, \emph{Musical}, \emph{Science Fiction}.

\par \emph{Ohsumed}~\cite{hersh1994ohsumed}: This dataset is a subset of the MEDLINE database. The labels correspond to 23 Medical Subject Headings categories of cardiovascular diseases group.

\par \emph{TMC2007}\footnote{https://c3.nasa.gov/dashlink/resources/138/}: The documents are the aviation safety reports corresponding to the one or more problems that occurred during certain flights. There are a total of 22 unique labels. 

\par \emph{Patents}\footnote{http://www.patentsview.org/download/}: This dataset contains summary text of the patents and the labels are the associated Cooperative Patent Classification (CPC) group labels. We randomly take a subset corresponding to the eight CPC groups: \emph{A: Human Necessities}, \emph{B: Operations and Transport}, \emph{C: Chemistry and Metallurgy}, \emph{D: Textiles}, \emph{E: Fixed Constructions}, \emph{F: Mechanical Engineering}, \emph{G: Physics}, \emph{H: Electricity}.

\par \emph{Delicious}~\cite{zubiaga2009content}: This data set contains tagged web pages retrieved from the website delicious.com.
We randomly choose documents corresponding to 20 common tags: \emph{humour}, \emph{computer}, \emph{money}, \emph{news}, \emph{music}, \emph{shopping}, \emph{games}, \emph{science}, \emph{history}, \emph{politics}, \emph{lifehacks}, \emph{recipes}, \emph{health}, \emph{travel}, \emph{math}, \emph{movies}, \emph{economics}, \emph{psychology}, \emph{government}, \emph{journalism}.

All datasets were pre-processed by removing stop-words and rare words (words that occur less than 4 times), by stemming the words with Porter stemmer~\cite{porter1980algorithm}, and by removing sentences with less than 4 or more than 10 words. 

\subsection{Evaluation Methodology and Performance Assessment}
\subsubsection{Methodology}
To evaluate our method on the segmentation task, we require ground-truth data for the segmentation and associated labels. Since we are not aware of any publicly available datasets with segment-level labels, we created a synthetic test set to evaluate our methods on the segmentation task as follows: For each of the multilabel datasets described in the previous section, first we separated the single-label documents from the other (containing at least two labels) documents. From this set of single-label documents, we randomly chose documents and merged them to form a multilabel document. To mimic the naturally co-occurring label sets, we only created the documents whose resultant label-set is present in the naturally occurring multilabel documents. Since we generated these documents synthetically, we know the segments and the associated labels. This allows us to assess the performance of our methods on the segmentation task. To predict the segmentation of a document in this test set, we consider the label set as the set of all classes ($L_d = C$). We used the multilabel documents (documents with at least two labels) to create the training and test set to test our methods on the multilabel classification task. We did an 80--20 split to form the training-test sets. We used the same training set to estimate the classifier required for segmentation.

\subsubsection{Metrics}
For evaluation on the segmentation task, we first look into per-point prediction accuracy, which corresponds to the fraction of sentences that are predicted correctly. We call this metric $\pppa$ (per-point prediction accuracy) and define it as:

\begin{equation}
    \pppa(S_1, S_2) = \frac{1}{N}\sum_{i=1}^N 1(S_1(i) == S_2(i)).
\end{equation}

As a single-point measure, $\pppa$ does not take into account the correlation between the neighboring sentences. To overcome this limitation of $\pppa$, we also report results on the Segment OVerlap score ($\sov$), which is a commonly used evaluation metric for protein secondary structure predictions~\cite{rost1994redefining}. $\sov$ measures how well the observed and the predicted segments align with each other. $\sov$ is defined as
\begin{equation}
    \sov(S_1, S_2) = \frac{1}{N}\sum_{\mathclap{ \substack{s_1\in S_1\\s_2 \in S_2\\(s_1, s_2) \in s}}}\frac{\minov(s_1, s_2) + \delta(s_1, s_2)}{\maxov(s_1, s_2)}\times \len(s_1),
\end{equation}
where $N$ is the total number of sentences in the document we are segmenting, $S_1$ is the observed segmentation, and $S_2$ is the predicted segmentation. The sum is taken over all segment pairs $(s_1, s_2) \in s$ for which $s_1$ and $s_2$ overlap on at least one point. 
\begin{wrapfigure}{r}{0.50\columnwidth}
\small
\centering
\centerline{\fbox{\includegraphics[width=1.7in]{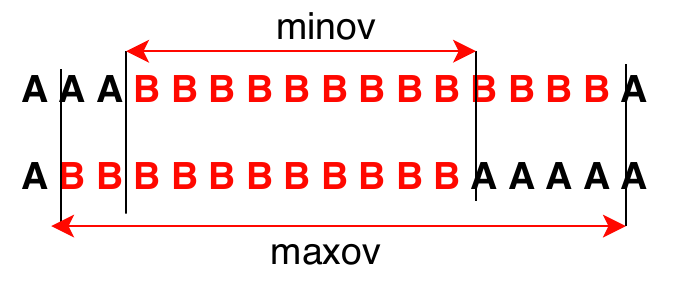}}}
\caption{\footnotesize Illustration of $\minov$ and $\maxov$. The segments compared are the ones labeled with letter \emph{B}.}
\label{minov}
\end{wrapfigure} 
The actual overlap between the $s_1$ and $s_2$ is $\minov(s_1, s_2)$, that is, the number of points both segments have in common, while $\maxov(s_1, s_2)$ is the total extent of both segments. Calculation of the $\minov$ and $\maxov$ is illustrated in Figure \ref{minov}. The accepted variation $\delta(s_1, s_2)$ is to bring robustness in case of minor deviations at the ends of segments. $\delta(s_1, s_2)$ is defined as~\cite{zemla1999modified}:
\begin{equation}
    \delta(s_1, s_2)=\min
    \begin{cases}
      \maxov(s_1, s_2) - \minov(s_1, s_2). \\
      \hfil \minov(s_1, s_2). \\
      \hfil \lfloor\len(s_1)/2\rfloor. \\
      \hfil \lfloor\len(s_2)/2\rfloor.
    \end{cases}
\end{equation}

Compared to the $\pppa$ metric, $\sov$ penalizes fragmented segments and favors continuity in the predictions. For example, prediction errors at the end of segments will be penalized less by $\sov$ than the prediction errors in the middle of the segments.

In addition to performing segmentation, we also used our methods to solve the multilabel classification problem. This was done by predicting for each document the union of the labels that were identified during the segmentation process. To evaluate the performance of our methods on this task, we used the following measures: The first is the $\ef$ score. For each document, we compute the $\ef$ score based on the predicted label set and the observed label set. We report the mean $\ef$ score ($\ef_{mean}$) over all the documents. $\ef_{mean}$ is an important metric because it tells how well the predicted labels of the sentences correspond to the document labels. 

The second is the Area Under the Receiver Operating Characteristic Curve ($\auroc$)~\cite{bradley1997use}. $\auroc$ gives the probability that a randomly chosen positive example has a higher probability of being positive than a randomly chosen negative example. We report $\auroc$ both under the micro ($\auroc_{\mu}$) and macro ($\auroc_{M}$) settings. $\auroc_{M}$ treats all the classes equally, whereas $\auroc_{\mu}$ aggregates the contributions of all classes to compute the metric. To compute $\auroc$ metrics, we need soft scores corresponding to the likelihood of belonging to a class. We find the score of a document corresponding to a class as the maximum of the scores of any segment of that document for that class.

\subsection{Parameter selection}
We used a two-layer perceptron for multiclass classification with 512 nodes in the hidden layer. For regularization, we used a dropout~\cite{srivastava2014dropout} of $0.5$ between the input and hidden layer. For optimization, we used ADAM~\cite{kingma2014adam}. For both SEG-NOISY and SEG-REFINE, we ran a total of three iterations. In each iteration, we trained the network for 100 epochs. We set the learning rate to 0.001 for initial training of the network and used 0.0001 as the learning rate for fine-tuning.

\begin{wraptable}{r}{0.65\columnwidth}
\footnotesize
  \caption{Hyperparameter values.}
  \begingroup
        \setlength{\tabcolsep}{1.5pt}
  \begin{tabularx}{2.2in}{lrrr}
    \hline
Dataset   & MLTM    & SEG-NOISY  & SEG-REFINE\\
          & $m$    & $\alpha$  & $\alpha$\\ \hline
Movies    & 120 & 0.30 & 0.55\\
Ohsumed   & 80 & 0.10 & 0.45\\
TMC2007   & 70 & 0.20 & 0.50\\
Patents   & 120 & 0.20 & 0.40\\
Delicious & 70 & 0.20 & 0.45\\ \hline
\end{tabularx}
\endgroup
  \label{tab:hyperparameters}
\end{wraptable}
We compare our approaches against MLTM.
For all methods, we chose the hyperparameters corresponding to the best performing model. The hyperparameter for our methods is $\alpha$ (new segment penalty) and the hyperparameter for the MLTM is $m$ (number of topics). The chosen values of these hyperparameters for different datasets are shown in Table \ref{tab:hyperparameters}.
\section{Results and Discussion}\label{results}

\subsection{Text segmentation}

\begin{table}[!t]
\footnotesize
\centering
  \caption{Results on the segmentation and classification tasks.}
     \begingroup
        \setlength{\tabcolsep}{3.0pt}
          \begin{tabularx}{\columnwidth}{Xlrrrrr}
            \hline
        Dataset   & Model    & $\sov$  & $\pppa$ & $\ef_{mean}$   & $\auroc_{\mu}$ & $\auroc_{M}$ \\ \hline
        Movies    & SEG-NOISY & 0.49 & 0.36  & 0.64 & \textbf{0.82}        & \textbf{0.81}        \\ 
        Movies    & SEG-REFINE & \textbf{0.51} & 0.38 & 0.62 & 0.81        & 0.79        \\ 
        Movies    & MLTM     & 0.50 & \textbf{0.40}    & \textbf{0.65} & \textbf{0.82}        & 0.80        \\ \hline
        Ohsumed   & SEG-NOISY & \textbf{0.63} & \textbf{0.48} & \textbf{0.64} & \textbf{0.94}        & \textbf{0.93}        \\ 
        Ohsumed   & SEG-REFINE & \textbf{0.63} & \textbf{0.48} & 0.63 & \textbf{0.94}        & 0.92        \\ 
        Ohsumed   & MLTM     & 0.55 & 0.46     & 0.59 & 0.93        & 0.90        \\ \hline
        TMC2007   & SEG-NOISY & 0.56 & 0.45 & \textbf{0.65} & \textbf{0.96}        & \textbf{0.92}        \\
        TMC2007   & SEG-REFINE & \textbf{0.59} & \textbf{0.46} & \textbf{0.65} & 0.95        & 0.89        \\ 
        TMC2007   & MLTM     & 0.50 & 0.44    & 0.62 & \textbf{0.96}        & 0.91        \\ \hline
        Patents   & SEG-NOISY & 0.53 & 0.42 & \textbf{0.63} & \textbf{0.87}        & \textbf{0.86}        \\ 
        Patents   & SEG-REFINE & \textbf{0.58} & 0.46 & 0.62 & 0.86        & 0.85        \\ 
        Patents   & MLTM     & 0.55 & \textbf{0.49}     & 0.59 & 0.85        & 0.84        \\ \hline
        Delicious & SEG-NOISY & 0.46 & 0.35 & 0.48 & \textbf{0.86}        & \textbf{0.85}        \\ 
        Delicious & SEG-REFINE & 0.48 & 0.35 & 0.48 & 0.84        & 0.84  \\
        Delicious & MLTM     & \textbf{0.49} & \textbf{0.37}    & \textbf{0.50} & 0.84        & 0.83        \\ \hline
        \end{tabularx}
        \endgroup
          \label{tab:segmentation}
\end{table}
\subsubsection{Quantitative evaluation}
The metrics $\sov$ and $\pppa$ in Table \ref{tab:segmentation} show the performance for various methods on the segmentation task. Except for the \emph{Ohsumed} dataset, SEG-REFINE performs better than the SEG-NOISY. On the \emph{Ohsumed} dataset, SEG-REFINE performs at par with SEG-NOISY. The average performance gain of SEG-REFINE over SEG-NOISY over all the datasets, in terms of $\sov$ metric is $4.5\%$. This illustrates that the segmentation can give us more precise examples of text segments belonging to a class, leading to better classification models. 

Except for the Delicious dataset, our methods perform better than the MLTM on the $\sov$ metric and give comparable performance on the Delicious dataset. The performance gain of SEG-REFINE over MLTM averaged over all the datasets, in terms of the $\sov$ metric is $7.7\%$. In contrast, on the $\pppa$ metric, MLTM shows better performance than our methods in three out of five datasets and almost equal performance on the other two datasets. Lower performance on the $\sov$ metric but better performance on the $\pppa$ metric shows that MLTM forms fragmented segments. On the other hand, our methods avoid fragmentation by assigning the neighboring sentences same labels, leading to better performance on the $\sov$ metric.

The somewhat lower performance of our methods on the $\sov$ metric with respect to the Delicious dataset can be explained through its class labels, that portray overlapping concepts. For example, the labels \emph{government} and \emph{politics} are very similar. Moreover, as the labels in this dataset are bookmarked by web users, the same document can be tagged as \emph{government} and \emph{politics} by different users, but it will be difficult to associate parts of the document to either of these labels. Therefore, our assumption that each segment can be mapped to one of the class labels of the document is difficult to hold in such cases. We provide further qualitative discussion regarding this in Section \ref{qualitative}. 

\subsubsection{Qualitative evaluation}\label{qualitative}
\begin{figure*}[!t]
\centering
\subfloat[Ground truth]{\includegraphics[width=0.32\textwidth]{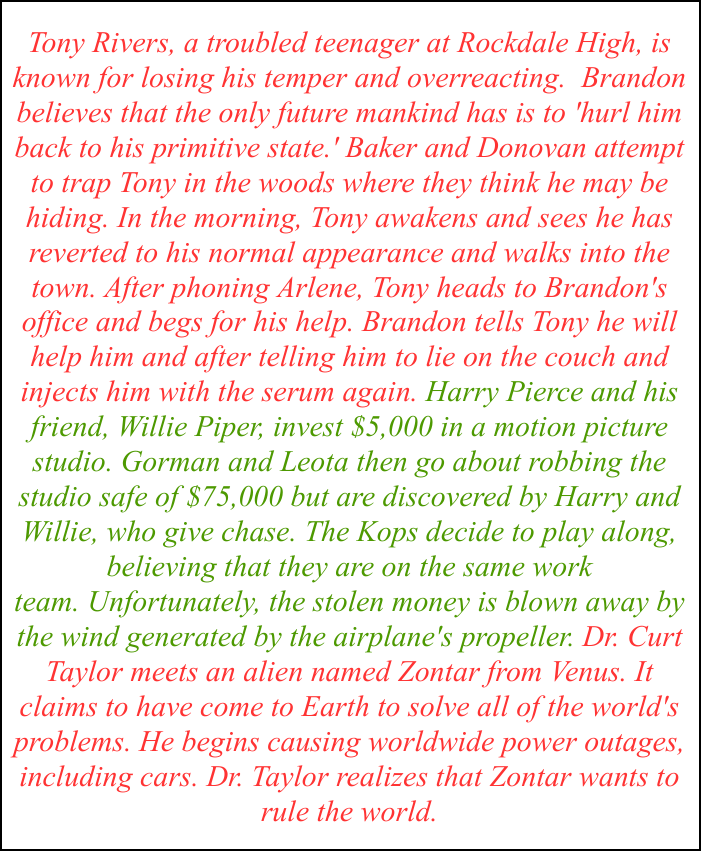}%
\label{fig:qual_ground}}
\hfil
\subfloat[SEG-REFINE]{\includegraphics[width=0.32\textwidth]{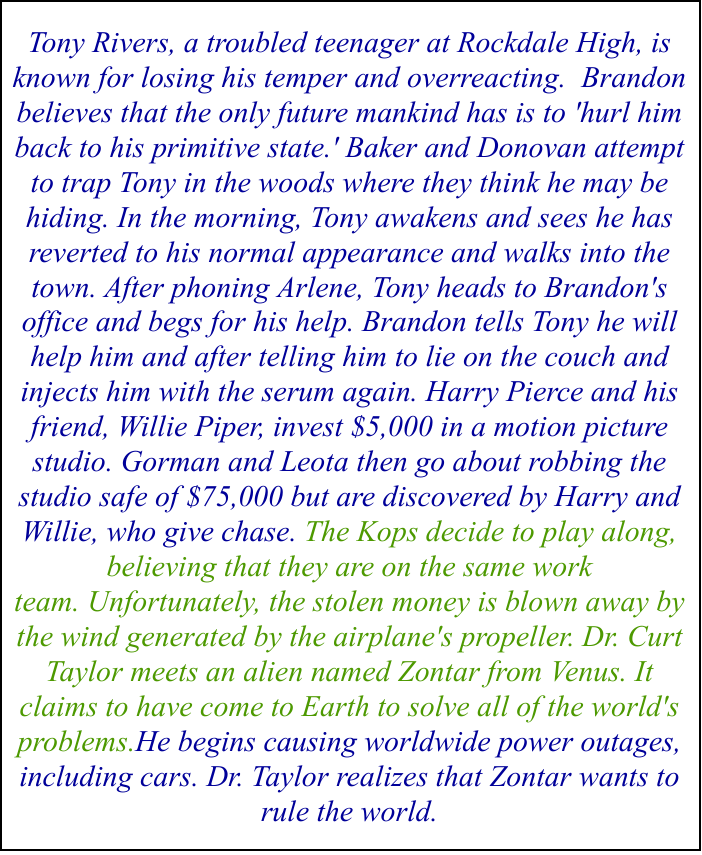}%
\label{fig:qual_seg_refine}}
\hfil
\subfloat[MLTM]{\includegraphics[width=0.32\textwidth]{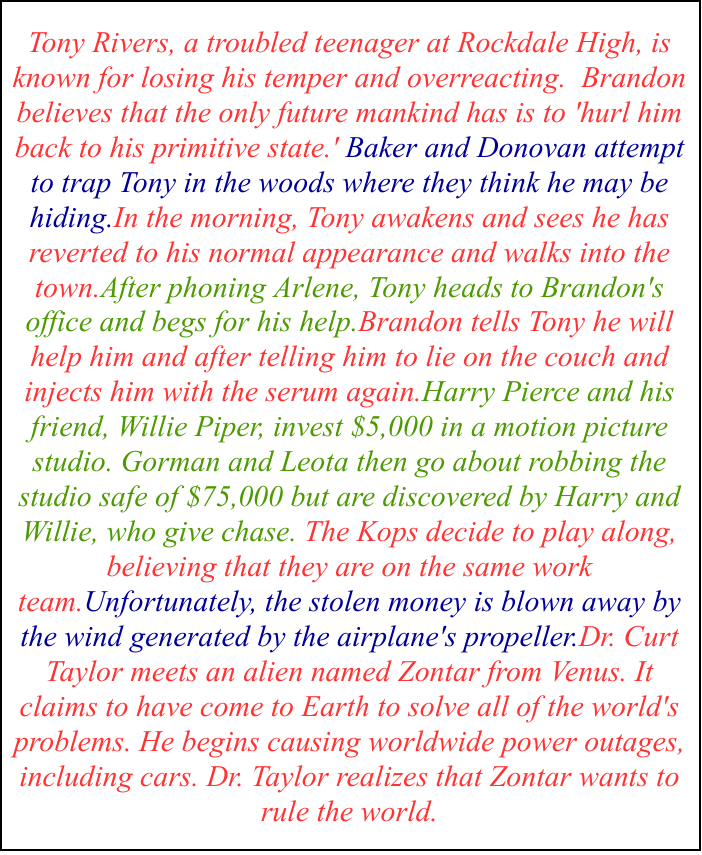}%
\label{fig:qual_mltm}}
\caption{Segmentation and annotation of a synthetically generated document from the Movies dataset. Color code: Red text: \emph{Science Fiction}, green text: \emph{Comedy} and blue color: \emph{Action}. MLTM generates fragmented segments and is able to annotate with both \emph{Action} and \emph{Science Fiction} genres. The segments generated by SEG-REFINE are not fragmented but the segments are annotated with \emph{Action} instead of \emph{Science Fiction} because of the overlapping content.}
\label{fig:qual}
\end{figure*}

In order to visualize the segmentation produced by our methods and compare them with the one produced by MLTM, we look into one of the documents that was synthetically generated from the Movies dataset. The labels associated with the document are \emph{Science Fiction} and \emph{Comedy}. Figure \ref{fig:qual_ground} shows the ground-truth segmentation and corresponding labels for the synthetically-generated document. The provided example is interesting as it also shows the case of overlapping labels, because the label \emph{Science Fiction} overlaps with other labels like \emph{Action} and \emph{Thriller}. Figure \ref{fig:qual_seg_refine} shows the corresponding segmentation generated by SEG-REFINE. SEG-REFINE is able to find contiguous and semantically-coherent segments, although instead of \emph{Science Fiction}, it labels the corresponding segments as \emph{Action} because of the overlapping nature of the \emph{Science Fiction} and \emph{Action} genres. 
Figure \ref{fig:qual_mltm} shows the segmentation generated by MLTM. MLTM is able to annotate segments with both the \emph{Action} and \emph{Science Fiction} genres. But, the segments produced by MLTM are fragmented, as MLTM makes the predictions for each sentence irrespective of the neighboring sentences.

\subsection{Multilabel classification}
The metrics $\ef_{mean}$, $\auroc_{\mu}$ and $\auroc_{M}$ in Table \ref{tab:segmentation} show the performance of different methods on the classification task. SEG-NOISY and SEG-REFINE perform similarly on all the three metrics. This indicates that SEG-NOISY is able to predict the labels in a document as well as SEG-REFINE. However, the locations of these predicted labels are not as precise as predicted by SEG-REFINE, which leads to at par performance on the multilabel classification task but lower performance on the segmentation task.

Similar to the segmentation task, except for the Movies and Delicious datasets, our methods perform better than the MLTM on the $\ef_{mean}$. This shows that the labels predicted for the sentences by our methods correlate better with the document labels as compared to the labels predicted by the MLTM for the Ohsumed, TMC2007 and Patents datasets. The average performance gain of SEG-REFINE over MLTM on these three datasets is $5.5\%$. The somewhat lower performance on the Movies and Delicious is again because of the overlapping labels, due to which the labels predicted for the sentences do not exactly match the document labels. 

With respect to the $\auroc_{\mu}$ and $\auroc_{M}$ metrics, our methods perform as good, if not somewhat better than the MLTM. 
Moreover, our approaches are resource-lean as they take an order of magnitude less time to execute as compared to the MLTM. For example, the complete time taken for training and testing on the Movies dataset is as follows: SEG-NOISY took 27 minutes, SEG-REFINE took 26 minutes and MLTM took more than 19 hours. The results correspond to experiments ran on the same machine. Another point to note is that our methods are dependent on the underlying multiclass classifier. The performance of our methods can potentially be improved by using better classification models like deeper networks and recurrent networks to capture the sequence information within the text.
\section{Conclusion}\label{conclusion}
In this paper, we proposed a method to automatically segment documents into semantically-coherent segments and annotate these segments with the corresponding class label. Our method improves upon the prior approaches by exploiting the structure within the documents. Experiments on the text segmentation task and multilabel classification task show the added advantage of using the proposed method. 

\section*{Acknowledgment}
This work was supported in part by NSF (1447788, 1704074, 1757916, 1834251), Army Research Office (W911NF1810344), Intel Software and Services Group, and the Digital Technology Center at the University of Minnesota. Access to research and computing facilities was provided by the Digital Technology Center and the Minnesota Supercomputing Institute.

\bibliographystyle{IEEEtran}  
\bibliography{refs}

\end{document}